\documentclass[runningheads]{llncs}
\usepackage[T1]{fontenc}
\usepackage{graphicx}
\begin{document}
\title{Emerging categories in scientific explanations}
%
%
\author{Giacomo Magnifico\orcidID{0000-0002-1782-0668} \and
Eduard Barbu\orcidID{0000-0002-3664-5367}} 
\authorrunning{G. Magnifico, E. Barbu}
%
\institute{Institute of Computer Science, University of Tartu, Tartu, Estonia\\
\email{\{name.surname\}@ut.ee}}

%
%

%
\maketitle              
\begin{abstract}
Clear and effective explanations are essential for human understanding and knowledge dissemination. The scope of scientific research aiming to understand the essence of explanations has recently expanded from the social sciences to include the fields of machine learning and artificial intelligence. Important contributions from social sciences include \cite{miller2019explanation,jsmill,thagard,lombrozo2006structure,halpern2005causes,lewis1986causal} with works that examine critical aspects such as causality (cause-and-effect relationships), contrast (distinctions between differing scenarios), relevance (applicability of explanations), and truth (accuracy and verifiability of explanations). However, machine learning and natural language processing focus more on operational definitions and on the importance of constructing datasets, as seen in studies by \cite{tan-2022-diversity,wiegreffe,hartmann}. Since explanations for machine learning decisions must be both impactful and human-like \cite{kulesza2015principles,ali2023explainable,mill2a,mill2b,mill2c}, a major challenge lies in developing explanations that emphasize proximal aspects — details that are immediately relevant, direct and related to the user — over broad algorithmic processes \cite{tan-2022-diversity}. The current lack of large-scale datasets with a focus on both human-like and human-generated explanations \cite{wiegreffe} highlights the issue addressed by this work.

The specific research questions of this work are thus the following: \textit{in what form(s) do explanations take form within the context of scientific literature? Can we provide an annotated dataset with as clear-cut definitions as possible and reach an acceptable consensus between different annotators?}

The scope of this study has been limited to \textbf{scientific literature} due to the intrinsic nature of explanations -- to avoid complications that would derive from the additional analysis of truth and relevance. Scientific explanations possess an identifiable general structure that involves a relationship between two components: the \textit{explanans} (which provides the explanation) and the \textit{explanandum} (what is being explained). The explanandum is contingent on the explanans, as changes in the latter directly impact the former. A useful example can be the equation \(y = 2*x\) ; the value of \(y\) (explanandum) depends on the value of \(x\) (explanans), as it increases with the increase of \(x\), but the inverse also holds true. This relationship shows how explanations build upon the dependence of the explanandum on the explanans while providing enough nuance for multiple types of explanations within these limitations. An additional constraint was to only include explanations that presented an \textbf{explicit explanandum}, e.g. \textit{“the sky is blue because of light refraction through the atmosphere”} rather than \textit{“this happens due to light refraction through the atmosphere”}, in order to avoid explanations trailing through multiple sentences.

With our research questions in mind, we started by extracting sentences that indicate explanations from scientific literature among various sources in the biotechnology and biophysics topic domains, the majority of which were selected from PubMed's PMC Open Access subset. The selected 340 sentences were then analyzed and different "explanation types", as possible categories for interpretation, emerged from the data. It’s crucial to reiterate that this categorization process was entirely driven by the dataset, in a deductive classification originating from the text and not a superimposition of pre-existing categories upon the dataset. This method, therefore, \textit{avoided pre-set criteria to explore instead the intrinsic connections between categories and the dataset’s subject}, aiming to understand the commonalities and differences within the explanations.
The categories of explanations that emerged are the following: \textbf{causation}, which establish a cause-and-effect relationship, stating that one event or condition leads to another without detailing intermediate steps \cite{Mackie1974-MACTCO-7}; \textbf{mechanistic causation}, which detail the underlying mechanisms by which a cause leads to an effect, outlining the intermediate steps that explain how and why the cause produces the outcome \cite{Machamer2000-MACTAM}; \textbf{contrastive}, which focus on comparing scenarios to explain why a particular outcome occurred in one case but not in another, emphasizing divergent outcomes \cite{DBLP:journals/corr/abs-2103-01378}; \textbf{correlation}, which detail relationships between variables where changes in one are associated with changes in another but without establishing causality; \textbf{functional}, which focus on the function of a trait in relation to its form and effectiveness, particularly in biology \cite{Mayr1988}; \textbf{pragmatic approach}, which focus on the selection of choices/actions based on convenience/effectiveness, requiring a conscious choice and emphasizing practicality or applicability \cite{Morgan1999-MORMAM-7}.

To minimize author bias in sentence categorization, we conducted a classification study on the Prolific platform \cite{prolific-citation} with 120 annotators divided between 10 questionnaires, guaranteeing a base of twelve annotators per sentence. Each annotator completed the questionnaire in one sitting, with a median completion time of 35 minutes, and was compensated at an average rate of £8/hour. After sanity checks and the removal of statistical outliers, 10 evaluations per sentence were kept along with the highest-quality 272 explanatory sentences. Upon calculating the averaged Krippendorf's alpha \cite{krippendorff-alpha-1,krippendorff-alpha-2} value to gauge the robustness of inter-annotator agreement, significant disagreement between categories of similar causal strength was observed (causation/mechanistic causation, correlation/functional/pragmatic approach). After categorizing the sentences by causal strength and the number of relations, with the new categories of \textbf{strong relation} (causation and mechanistic causation), \textbf{weak relation} (correlation, functional, pragmatic approach) and \textbf{multi-path relation} (contrastive), the average agreement between annotators improved to a value of \textbf{0.667} . Albeit only slightly over a desiderata target, the final alpha value is still a representation of good agreement between annotators and, thus, of a high-quality human-annotated explanation dataset. The dataset is made available to the community through a dedicated repository at \cite{github}.

\end{abstract}
%
%
%
%
%
%
 \bibliographystyle{splncs04}
 \bibliography{bibliography}
%
%
%
%
%
\end{document}